\documentclass{article}

\usepackage[nonatbib,preprint]{nips_2018}

\usepackage[utf8]{inputenc} 
\usepackage[T1]{fontenc}    
\usepackage{hyperref}       
\usepackage{url}            
\usepackage{booktabs}       
\usepackage{amsfonts}       
\usepackage{amsmath,amsthm,amssymb}
\usepackage{mathrsfs}
\usepackage{enumitem}
\usepackage{algorithm}
\usepackage{algorithmic}
\usepackage{nicefrac}       
\usepackage{microtype}      
\usepackage{mcr}
\usepackage{pifont}
\usepackage{graphicx}
\usepackage{subcaption}
\usepackage{epstopdf}

\newcommand{\R}{\mathbb{R}}

\newcommand{\mbI}{\mathbf{I}}
\newcommand{\mbP}{\mathbf{P}}

\newcommand{\mbR}{\mathbf{R}}

\newcommand{\mbPi}{\mathbf{\Pi}}

\newcommand{\mcA}{\mathcal{A}}

\newcommand{\mcS}{\mathcal{S}}

\title{Towards Generalized Inverse Reinforcement Learning}

\author{
	Chaosheng Dong\thanks{Work done prior to joining Amazon.} \\
	Department of Industrial Engineering\\
	University of Pittsburgh\\
	Pittsburgh, PA 15260 \\
	\texttt{chaosheng@pitt.edu} \\
	\And
        Yijia Wang \\
	Department of Industrial Engineering\\
	University of Pittsburgh\\
	Pittsburgh, PA 15260 \\
	\texttt{yiw94@pitt.edu} \\
}

\begin{document}
	
	\maketitle
	
	\begin{abstract}
		This paper studies generalized inverse reinforcement learning (GIRL) in Markov decision processes (MDPs), that is, the problem of learning the basic components of an MDP given observed behavior (policy) that might not be optimal. These components include not only the reward function and transition probability matrices, but also the action space and state space that are not exactly known but are known to belong to given uncertainty sets. We address two key challenges in GIRL: first, the need to quantify the discrepancy between the observed policy and the underlying optimal policy; second, the difficulty of mathematically characterizing the underlying optimal policy when the basic components of an MDP are unobservable or partially observable. Then, we propose the mathematical formulation for GIRL and develop a fast heuristic algorithm. Numerical results on both finite and infinite state problems show the merit of our formulation and algorithm.
	\end{abstract}

\section{Introduction}
	\label{sec: introduction}
	Understanding human or animal's intrinsic objective or goal is crucial for modeling their behaviors. Nevertheless, as in most scenarios, one can only observe their behaviors, while cannot directly access their objectives. To bridge the discrepancy, inverse reinforcement learning (IRL) has been proposed and received significant research attention, which aims to learn a reward function that renders the observed policy optimal \cite{russell1998learning,ng2000algorithms}. Extending from its initial form \cite{ng2000algorithms}, IRL has been further developed and applied to learn the reward functions in more complex and realistic situations \cite{abbeel2004apprenticeship,Syed:2008:ALU:1390156.1390286,ziebart2008maximum,ratliff2009learning,ho2016model}.
	
	Despite its successes in many applications, IRL has several issues. Actually, the first major issue has been mentioned as an open question remained to be addressed in \cite{ng2000algorithms}. That is, in real-world empirical applications of IRL, there might be substantial noise in the observer's measurements of the agent's actions; moreover, the agent's own action selection process might be noisy and/or suboptimal; worse still, the agent's behavior might be strongly inconsistent with optimality. Consequently, solving IRL with the assumption of the optimality of the observed policy might lead to drastically inconsistent estimations of the components. To answer this question, our key step is to design an appropriate metric to quantify the discrepancy between the observed policy and the underlying optimal policy. 
	
	The second issue has been more overlooked, and needs subtler tool to fix. Under the framework of IRL, the reward function is the only component that the observer seeks to infer from the observation of an agent's behavior. This precludes a variety of useful things the observer can learn. Indeed, the agent's transition probabilities matrices, actions, or states, might be unobservable or partially observable as well. Such information about an agent and her environment, if obtained, should also be of a great value to the observer in understanding the decision schemes of the agent. Consider, for example, a grid world with one blocked square whose position is uncertain. Suppose the observer knows that it is within a set of squares. Given the observed policy, her goal is to figure out the position of the blocked square. This learning task does not fit the standard IRL framework.
	
    Our starting point of generalized inverse reinforcement learning (GIRL) is not to consider inverse reinforcement learning as merely constructing the reward function, but rather to construct all the unknown components of a Markov decision process (MDP) (e.g. reward function, transition probability matrices, states, actions). Moreover, we combine the process of constructing the unknown components of an MDP with the process of reconstruction of the underlying optimal policy, and we argue that the two processes interact with each other in a mutually reinforcing way. In this paper, we address two inter-related objectives of GIRL: 1) to reconstruct the optimal policy from the observed policy; and 2) to find the unknown components of an MDP that can explain the reconstructed policy. The basic idea of our formulation for GIRL is to treat both the underlying optimal policy and the unknown components as random variables, and then reconstruct them by minimizing the discrepancy between the observed policy and the underlying optimal policy, subject to the constraints for the characterizations of the optimal policy.
	
	\textbf{Related works}\;\; Our idea of GIRL draws inspirations from two lines of research. The first line of research is mainly done in the machine learning community, which focuses on learning the reward function. There have been many exciting contributions to the IRL literature since the seminar work \cite{ng2000algorithms}. Some standouts are  \cite{ratliff2006maximum,ramachandran2007bayesian,ziebart2008maximum,NIPS2016_6420,pirotta2016inverse,finn2016guided,NIPS2017_6778,NIPS2017_6800}, which are developed to handle its limitations. Others are \cite{abbeel2004apprenticeship,Syed:2008:ALU:1390156.1390286,ziebart2008maximum,ratliff2009learning,ho2016model} which are applied to more complex and realistic situations in designing AI systems and modeling nature learning. Our paper studies the problem of learning not only the reward function, but also other basic components of an MDP which are unknown (e.g. transition probability matrices) or are only known to belong to an uncertainty set (e.g. action space, state space). 
	
	A parallel line of research occurs in the optimization and control community. Inverse optimization problem (IOP) is investigated in the papers \cite{ahuja2001inverse,dong2018ioponline,dong2018inferring,dong2020imop,dong2021wasserstein-aaai,dong2023imoponline,yu2023learning}. Here the goal is to find a minimal adjustment of the cost function such that the given solution becomes optimal for a linear programming problem. Interestingly, it shows that their IOP using $L_{1}$ or $ L_{\infty} $ norm to measure the distance between the estimated cost function and the given one is also a linear program, which coincides with the formulation of IRL in \cite{ng2000algorithms}. Inverse optimal control (IOC) is studied in \cite{boyd1994linear,keshavarz2011imputing,Nguyen2017}. The goal of this type of inverse problems is to construct a cost function or a feasible set such that the optimal solution of their associated optimization problem is equivalent to the given control law. 
	
\begin{figure}[H]
		\centering
		\includegraphics[width=0.9\linewidth]{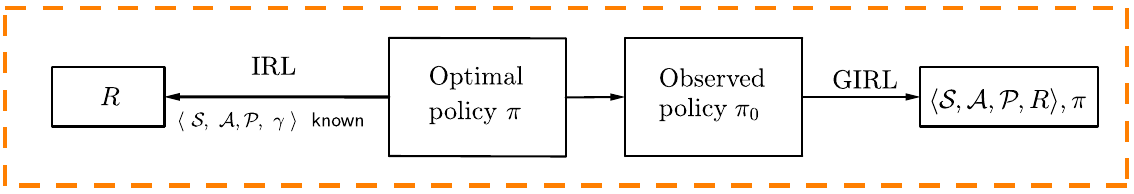}
		\caption{IRL stands for the inverse reinforcement learning. GIRL stands for the generalized inverse reinforcement learning studied in our paper. The observed policy $ \pi_{0} $ might be non-optimal.}
		\label{fig:difference between IRL and GIRL}
	\end{figure}

	\textbf{Our contributions}\;\;
	In Section 3, we propose the concept of policy matrix whose element represents the probability of taking an action in each state under the corresponding policy. Based on this new concept, we propose a well defined metric that quantifies the difference between two policies. Leveraging the proposed policy matrix and distance between policies, we present in Section 4 the mathematical formulation for GIRL in finite state spaces. An overview of the difference between IRL and GIRL is given in Figure \ref{fig:difference between IRL and GIRL}. We then customize GIRL by adding some constraints or penalties for different learning tasks. We propose a global search algorithm to solve GIRL and extend GIRL to large state spaces. Finally, we illustrate in Section 5 the performance of the algorithm on both a discrete grid world problem and a continuous grid world problem. Preliminary results show that our algorithm can solve GIRL quite efficiently and has a high quality inference of the unknown components of an MDP.
	\vspace{-1em}
\vspace{-3em}

\section{Problem Settings}
	\label{sec: problem settings}
	We introduce some notations and definitions for MDPs, reinforcement learning (RL), and IRL in this section. 
	Before moving on, we fix some useful notations. Let $ [s] = \{1,\ldots,s\} $ for any natural number $ s $, let $ \mathbf{1}^{s} = (1,\ldots,1)^{T} $ and $ \zero^{s} = (0,\ldots,0)^{T} $ denote all-ones vector and all-zeros vector in $ \bR^{s} $ respectively. Denote $ \norm[F]{\cdot} $ the Frobenius norm of a matrix. Let $ P(s) $ be the s-th row of matrix $ P $. The symbols $ \succeq $ and $ \succ $ denote non-strict and strict vectorial inequality respectively. 
	
\subsection{Markov Decision Processes} \label{sec: MDP}
	An MDP is specified by its state space $ \mcS $, action space $ \mcA $, transition function $ P : \mcS \times \mcA \rightarrow \mcS $, reward function $ R: \mcS \times \mcA \rightarrow \R $ with upper bound $ R_{\text{max}} $, initial state distribution $ D $ (it is suppressed for notational brevity when we refer to an MDP), and discount factor $ \gamma \in [0,1) $. In this paper, we write $ R(s) $ rather than $ R(s,a) $ for convenience. 
	
	
	\begin{definition}[Policy] \label{def. policy}
		A \textbf{policy} $ \pi(a|s) $ is the probability of taking action $ a $ in state $ s $. Denote $ \pi(s) $ the vector of probabilities of taking each action in state $ s $. 
	\end{definition}
	\begin{remark} \label{rmk. policy}
		\emph{(i)} Specially, when $ \pi $ is a \textbf{deterministic policy}, $ \pi(s) $ indicates a deterministic action in state $ s $. 
		\emph{(ii)} There exists at least one optimal policy that is stationary and deterministic for every MDP \cite{littman1994markov}. Throughout this paper we consider only the optimal policy that is stationary and deterministic.
    \end{remark}

	\begin{definition}[Transition probability matrix]
		\emph{(i)} $ \mbP^a $ is the transition probability matrix upon taking action $ a $, and $ P_{ij}^a $ is its $ (i,j) $th element, specifying the transition probability from state $ i $ to state $ j $ upon taking action $ a $. 
		\emph{(ii)} $ \mbP^\pi $ is the transition probability matrix under policy $ \pi $. For a deterministic policy, $ \mbP^\pi(s) $ indicates the row of transition probabilities in state $ s $ under policy $ \pi $, i.e. $ \mbP^\pi(s) = \bigl(P^{a_1}_{s,1}, P^{a_1}_{s,2}, \ldots, P^{a_1}_{s,|S|}\bigr) $, $ \pi(a_1|s) = 1 $.
	\end{definition}
	 
	

\subsection{Reinforcement learning and inverse reinforcement learning} \label{sec: IRL}
   
    RL methods seek the optimal policy basing on feedbacks (reward) from the environment. Specifically, given the state space $ \mcS $, action space $ \mcA $, reward function $ R $, and discount factor $ \gamma $, RL methods help to find an optimal policy $ \pi $ without knowledge about the transition probability matrices. There are many algorithms for RL, for example SARSA \cite{rummery1994line,singh1996reinforcement}, Q-learning \cite{watkins1989learning}, actor-critic methods \cite{werbos1974beyond,witten1977adaptive}, policy gradient methods \cite{glynn1987likelilood,sutton2000policy}, et. al.
	
	In contrast, IRL aims to find a reward function that can explain the observed behavior of an agent. 
	The basis of IRL is Proposition \ref{prop:optimal condition}.

	\begin{proposition}\label{prop:optimal condition}
		Given a finite state space $ \mcS $, a finite action space $ \mcA $, transition probability matrices $ P $, and discount factor $ \gamma \in (0,1) $, a deterministic policy $ \pi $ is optimal if and only if the reward $ \mbR $ satisfies
		\begin{align}\label{optimal condition}
		\tag{Optimal condition}
			(\mbP^{\pi} - \mbP^a) (\mbI - \gamma \mbP^{\pi})^{-1} \mbR \succeq \zero^{|\mcS|} \;\;\;\; \forall a \in \mcA
		\end{align}
	\end{proposition}

	The mathematical formulation for an IRL problem provided by \cite{ng2000algorithms} is 
	\begin{align}
	\max\;\;\;\; & \sum_{s\in \mcS} \min_{a \in \mcA\setminus\{\pi(s)\}} \{(\mbP^{\pi}(s) - \mbP^a(s)) (\mbI - \gamma \mbP^{\pi})^{-1} \mbR\} - \lambda \|\mbR\|_1 \label{eq. IRL objective} \\
	\text{s.t.}\;\;\;\; 
	& (\mbP^{\pi} - \mbP^a) (\mbI - \gamma \mbP^{\pi})^{-1} \mbR \succeq \zero^{|\mcS|} &  \forall a \in \mcA \label{eq. IRL constraint}\\
	& |\mbR(s)| \leq R_{\text{max}} &  \forall s \in \mcS \label{eq. IRL constraint2}
	\end{align}
	where constraints \eqref{eq. IRL constraint} describe the necessary and sufficient conditions for $ \pi $ being an optimal policy. Constraints \eqref{eq. IRL constraint} hold strictly if and only if $ \pi $ is the unique optimal policy. Constraints \eqref{eq. IRL constraint2} guarantee a finite solution. In most problems, there are multiple choices of $ \mbR $ that makes policy $ \pi $ optimal. To select one out of these choices, \eqref{eq. IRL objective} considers to infer a sparse $ \mbR $ that maximizes the difference between the quality of the optimal action and the quality of the next-best action.

\section{Policy matrix and distance between policies}\label{sec: definitions}
In this section, we propose the concept of policy matrix, whose element represents the probability of taking an action in a state under the corresponding policy. Based on this new concept, we propose a distance metric to quantify the difference between two policies. 
	
	\begin{definition}[Policy matrix]
		A policy $ \pi $ can be described by an $ |\mcS| \times |\mcA| $ matrix $ \mbPi $ such that 
		\begin{equation*}
		\mbPi(s,a) = \pi(a|s), \;\; \forall  s \in \mcS, \forall a \in \mcA
		\end{equation*}
		Matrix $ \mbPi $ is the policy matrix of policy $ \pi $.
		Specially, when the policy is deterministic, 
		\begin{align} \label{eq. policy matrix: deterministic}
			\mbPi(s,a) =
			\begin{cases}
			1 & \text{if} \; \pi(a|s) = 1 \vspace{1mm}\\
			0 & \text{if} \; \pi(a|s) \neq 1
			\end{cases}
		\end{align}
	\end{definition}

    \begin{remark}
    	\emph{(i)} The policy matrix $ \mbPi $ describes the probabilities of actions in each state under policy $ \pi $. 
    	\emph{(ii)} By the definition of policy, $ \sum_{a\in\mcA} \pi(a|s) = 1 \; \forall s \in \mcS$. Thus, $ \mbPi \,\one^{|\mcA|} = \one^{|\mcS|} $.
    	\emph{(iii)} When $ \pi $ is deterministic, only one element of each row of $ \mbPi $ equals to 1, and all others are 0. 
    	\emph{(iv)} According to Remark \ref{rmk. policy}, \eqref{eq. policy matrix: deterministic} holds for optimal policy. 
    \end{remark}
	
	\begin{definition}[Distance between policies]
		Let $ \pi_1, \pi_2 $ be two policies for an MDP $ \langle \mcS, \mcA, P, R, \gamma \rangle $. We define the distance between $ \pi_1 $ and $ \pi_2 $ as
			\begin{align*}
			d(\pi_1,\pi_2) = \norm[F]{\mbPi_{1} - \mbPi_{2}}
			\end{align*}
	\end{definition}
    Intuitively, $ d(\pi_1, \pi_2) $ measures the difference between the policy matrices of $ \pi_1 $ and $ \pi_2 $. It provides us a measure to quantify the difference between two policies. Two policies with larger distance are more different than those two with smaller distance.
    

    \begin{remark}
    	\emph{(i)} $ d(\pi_1, \pi_2) $ is a well-defined metric. The proof is given in supplementary material. \emph{(ii)} $ d(\pi_1,\pi_{2}) $ is similar to the Hamming Distance when $ \pi_1 $ and $\pi_{2} $ are both deterministic, which is the number of elements that the two policy matrices differ.
    	\emph{(iii)} In general, any matrix norm can be used to define the distance between policies, for example, $ l_1 $ norm, $ l_\infty $ norm. Throughout this paper we will focus on the Frobenius norm.
    \end{remark}
    
    \begin{theorem} \label{thm. optimal}
    	Given an MDP $ \langle \mcS, \mcA, P, R, \gamma \rangle $, a policy $ \pi $ is optimal if and only if its policy matrix $ \mbPi $ satisfies
    	\begin{align}
    	& (\mbP^{\pi} - \mbP^a) (\mbI - \gamma \mbP^{\pi})^{-1} \mbR \succeq \zero^{|\mcS|} && \forall a \in \mcA \vspace{1mm} \label{eq. constraint 1}\\
    	& \mbP^{\pi}(s) = \textstyle\sum\limits_{a \in \mcA} \mbP^a(s) \, \mbPi(s, a) && \forall s \in \mcS \vspace{1mm} \label{eq. constraint 2}
    	\end{align}
    	The proof is given in supplementary material. Constraints \eqref{eq. constraint 1} describe the characterizations of an optimal policy as stated in Proposition \ref{prop:optimal condition}. Constraints \eqref{eq. constraint 2} ensure that $ \mbP^{\pi}(s) $, the row vector of transition probabilities in state $ s $ under policy $ \pi $, is an element of the set $ \{\mbP^1(s), \mbP^2(s), \ldots, \mbP^{|\mcA|}(s)\} $.
    \end{theorem}
    
    \begin{remark}
    	Essentially, we rephrase the \ref{optimal condition} for a policy by using policy matrix. This theorem is established on the following important observations. The binary variable $ \mbPi(s, a) $ indicates whether action $ a $ shall be taken in state $ s $ under $ \pi $, one means yes, otherwise no. The transition probabilities in each state under $ \pi $ then can be expressed as the aggregation of the products of the indicator variable and the transition probabilities.
    \end{remark}

\section{Generalized inverse reinforcement learning}
	\label{sec: GIRL}
	Leveraging the proposed policy matrix, distance between policies, and Theorem \ref{thm. optimal}, we present in this section the mathematical formulation of GIRL for an MDP with finite state space. We then customize GIRL by adding constraints or penalties for different learning tasks. Note that GIRL has binary variables and thus is a non-convex problem. We propose a global search algorithm that can handle the non-convexity issue. Finally, we extend GIRL to large state spaces.
	
	\subsection{GIRL in finite state spaces} 
	\label{sec: GIRL in finite state spaces}
	Denote $ \pi_0 $ the observed policy that might carry measurement error or is generated with a bounded rationality of an agent, i.e., being suboptimal. Given $ \pi_{0} $, our aim is to learn not only a reward function and transition probability matrices, but also the action space and state space that are not exactly known but are known to belong to given uncertainty sets.

	
	We use $ \pi $ to denote the policy to be recovered.	Let $ \mbPi \in \{0,1\}^{|\mcS| \times |\mcA|} $ be the policy matrix of $ \pi $. Then, GIRL for the MDP $ \langle \mcS, \mcA, P, R, \gamma \rangle $ can be formulated as follows:
	
	\begin{align}
	\min\limits_{\mbPi} \;\;\;\; & \norm[F]{\mbPi - \mbPi_0} \vspace{1mm} \label{eq. GIRL objective}\\
	\text{s.t.} \;\;\;\; 
	& (\mbP^{\pi} - \mbP^a) (\mbI - \gamma \mbP^{\pi})^{-1} \mbR \succeq \zero^{|\mcS|} & & \forall a \in \mcA \vspace{1mm} \label{eq. GIRL constraint 1}\\
	& \mbP^{\pi}(s) = \textstyle\sum\limits_{a \in \mcA} \mbP^a(s) \, \mbPi(s, a) & & \forall s \in \mcS \vspace{1mm} \label{eq. GIRL constraint 2}\\
	& \mbPi \,\mathbf{1}^{|\mcA|} = \mathbf{1}^{|\mcS|} & & \label{eq. GIRL constraint 3}\\
	& \mbPi \in \{0,1\}^{|\mcS| \times |\mcA|} \label{eq. GIRL constraint 5}
	\end{align}
	where $ \mbPi_0 $ is the policy matrix of $ \pi_0 $. Constraints \eqref{eq. GIRL constraint 1} - \eqref{eq. GIRL constraint 2} have been explained in Theorem \ref{thm. optimal}. Constraints \eqref{eq. GIRL constraint 3}- \eqref{eq. GIRL constraint 5} imply that only one entry of each row of the policy matrix $ \mbPi $ is nonzero, and its value is forced to be 1. 
	The objective function \eqref{eq. GIRL objective} intends to minimize the difference between policy $ \pi $ and the observed policy $ \pi_0 $. Their distance is $ \norm[F]{\mbPi - \mbPi_0} $.

%

	
\subsection{Customizing GIRL for different learning tasks}\label{section:custimize girl}

\begin{table}[ht]
	\centering
	\caption{Extra constraints or penalty terms added to GIRL for different learning tasks}
	\label{table:extra constraints}
	\begin{tabular}{|c|c|c|c|}
		\hline
		\multicolumn{1}{|c|}{Reward} & \multicolumn{1}{c|}{Transition matrix} & \multicolumn{1}{c|}{Action} & \multicolumn{1}{c|}{State}\\ 
		\hline
		$\norm[\infty]{R} \leq R_{max}$, & $\mbP^{a_1} \succeq 0$, & $\mathbf{T} \in \{0,1\}^{|\tilde{\mcA}|}$, & $\mathbf{T} \in \{0,1\}^{|\tilde{\mcS}|}$, \\
		$\lambda\norm[0]{R} $ & $\mbP^{a_1} \,\one^{|\mcS|} = \one^{|\mcS|}$ & $\sum_{a \in \tilde{\mcA}}\mathbf{T}(a) = 1$, & $\sum_{s \in \tilde{\mcS}} \mathbf{T}(s) = 1$, \\
		or $\lambda\norm[1]{R} $ & & $\mbP^{a_{1}} = \sum_{a \in \tilde{\mcA}} \mathbf{T}(a) \mbP^{a}$ & $\mbP^{a} = \sum_{s \in \tilde{\mcS}} \mathbf{T}(s) \tilde{\mbP}^{a,s}, \; \forall a \in \mcA$ \\ 
		\hline
	\end{tabular}
\end{table}
We make the following assumption when $ \mcS $ or $ \mcA $ is known to belong to an uncertainty set. 
\begin{assumption} \label{assumption. state space and action space}
	The state space can be expressed as the union of two sets: $ \mcS = \mcS^\circ \cup \mcS' $, where $ \mcS^\circ $ is the set of observable states, $ \mcS' $ is the set of unobservable states. When the state space $ \mcS $ is not exactly known but is known to belong to a given uncertainty set $ \tilde{\mcS} $, we have $ \mcS' \neq \emptyset $, and $ \mcS' \subseteq \tilde{\mcS} $. 		
	The action space has the similar setting. 
\end{assumption}

Some constraints or penalty terms are added to GIRL when dealing with different learning problems.

\textbf{Reward function }
To ensure a finite reward function, $\norm[\infty]{R} \leq R_{max}$ is a standard constraint \cite{ng2000algorithms}.
Note that there might exist multiple reward functions in the optimal set for GIRL. We say $ R $ is non-identifiable in this situation. 
To find a proper $ R $, we can add a penalty to the objective function, for example $\lambda\norm[0]{R}$ or $\lambda\norm[1]{R} $ with $ \lambda\geq 0 $ is often used as a penalty term to infer a reward function that preserves some type of sparsity in statistics and machine learning literature \cite{tibshirani1996regression}. 
Then, this constraint and penalty listed in Table \ref{table:extra constraints} are added to GIRL.

\textbf{Transition probabilities matrices }
Without loss of generality, suppose the transition probabilities under action $ a_1 $, i.e. the transition probability matrix $ \mbP^{a_1} $ is partially observable. To guarantee the matrix we learn is a transition probability matrix, we need all elements to be positive, and elements of each row to sum up to one. 
Then, the two constraints listed in Table \ref{table:extra constraints} are added to ensure that $ \mbP_{a_1} $ is a properly defined transition probability matrix. 

\textbf{Actions }
Without loss of generality, suppose $ a_1 $ is unobservable but is known to belong to $ \tilde{\mcA} $. Let $ \mathbf{T}(a) \in \{0,1\} $ indicate whether $ a \in \tilde{\mcA} $ is the unobservable action. Let $\sum_{a \in \tilde{\mcA}}\mathbf{T}(a) = 1$ to ensure that exact one action in $ \tilde{\mcA} $ is the unobservable one. Let $\mbP^{a_{1}} = \sum_{a \in \tilde{\mcA}} \mathbf{T}(a) \mbP^{a}$ to ensure $ \mbP^{a_{1}} \in \{\mbP^{a} | a \in \tilde{\mcA} \} $. 
Then, the three constraints listed in Table \ref{table:extra constraints} are added to GIRL.

\textbf{States }
Without loss of generality, suppose $ s_1 $ is unobservable but is known to belong to $ \tilde{\mcS} $. Let $ \mathbf{T}(s) \in \{0,1\} $ indicates whether $ s \in \tilde{\mcS} $ is the unobservable state. Let $\sum_{s \in \tilde{\mcS}}\mathbf{T}(s) = 1$ to ensure that exact one state in set $ \tilde{\mcS} $ is the unobservable one. Denote $ \tilde{\mbP}^{a,s} $ the ``fake'' transition probability matrix under action $ a $ when $ s\in\tilde{\mcS} $ is the unobservable state. For each $ a \in \mcA $, let $\mbP^{a} = \sum_{s \in \tilde{\mcS}} \mathbf{T}(s) \tilde{\mbP}^{a,s} $ to ensure $ \mbP^a \in \{\tilde{\mbP}^{a,s} | s\in\tilde{\mcS}\} $. Then, the three constraints listed in Table \ref{table:extra constraints} are added to GIRL.

\textbf{Jointly learning} When more than one component of an MDP, say the reward function and a transition probability matrix, are unknown, we can simply customize GIRL by adding the extra constraints and penalties for both the reward function and the transition probability matrices in Table \ref{table:extra constraints}. 

\begin{remark}
	\emph{(i)} Although the additional constraints or penalties provided in Table \ref{table:extra constraints} are only for learning one unobservable transition probability matrix, action, or state, extensions to multiple ones are indeed not difficult. For example, suppose multiple transition probability matrices are unknown. Then we add multiple sets of the constraints as that in Table \ref{table:extra constraints}. \emph{(ii)} The key point in learning the unobservable action or state is to represent a transition probability matrix as the sum of all the candidates' transition probability matrices. To simplify the formulation, we suppose these candidates' transition probability matrices are available. Otherwise, it is a more complex jointly learning problem. 
\end{remark}

\subsection{An algorithm for solving GIRL and extensions to large state space}
	Note that we suppose the observed policy $ \pi_{0} $ might carry noise. If the action taken in a state under $ \pi_{0} $ is not optimal, then we call that state a \textbf{noisy state}. Specifically,  we can set an upper bound $ \eta $ for the number of noises in $ \pi_{0} $, and call it \textbf{noise threshold}, to indicate the noise level we believe in $ \pi_{0} $. Here, $ \eta \in \{0,\ldots,|\mcS|\} $. The larger $ \eta $ is, the more noises one believes $ \pi_{0} $  carries. Since GIRL is a nonconvex program, solving it exactly is practically infeasible for problem with large state space.  We thus propose a heuristic algorithm to handle the challenges in solving GIRL. 
	\begin{algorithm}[ht]
		\caption{Global search Algorithm for GIRL}
		\label{alg:online-iop}
		\begin{algorithmic}[1]
			\STATE {\bfseries Input:} Policy $ \pi_{0} $, components of $ \langle \mcS,\mcA,P,R,\gamma\rangle $ that are given, noise threshold $ \eta $.
			\STATE {\bfseries Initialize} Policy matrix $ \mbPi_{0} $, some large number Obj, and a guess of the unknown components $ \hat{\theta} $.
			\FOR{$ i_{1} = 1 $ to $ |\mcS| - \eta + 1 $, $ \ldots $, $ i_{\eta} = i_{\eta - 1} + 1 $ to $ |\mcS| $}
			\FOR{$ j_{1} = 1$ to $ |\mcA| $, $ \ldots $, $ j_{\eta} = 1$ to $ |\mcA| $}
			\STATE Let $ \mbPi(i,j) = \begin{cases}
			1, & (i,j) \in \{(i_{1},j_{1}),\ldots,(i_{\eta},j_{\eta})\} \\
			0, & i \in \{i_{1},\ldots,i_{\eta}\} \; \text{and}\;  (i,j) \notin \{(i_{1},j_{1}),\ldots,(i_{\eta},j_{\eta})\} \\
			\mbPi_{0}(i,j), & \text{otherwise}
			\end{cases} $
			\STATE Add specific constraints or penalty term listed in Table \ref{table:extra constraints}. Solve \eqref{eq. GIRL objective} - \eqref{eq. GIRL constraint 5} and get an estimation $ \theta $ of the unknown components.
			\IF{$\norm[F]{\mbPi - \mbPi_{0}} \leq \text{Obj} $ and GIRL has optimal solution in the previous step 6}
			\STATE Obj $ \leftarrow \norm[F]{\mbPi - \mbPi_{0}} $, $ \hat{\theta} \leftarrow \theta $ 
			\ENDIF
			\ENDFOR
			\ENDFOR
		\end{algorithmic}
	\end{algorithm}
\begin{remark}
	\emph{(i)} When a penalty term is added to GIRL, step 7 and 8 will be modified by replacing $ \norm[F]{\mbPi - \mbPi_{0}} $ with $\norm[F]{\mbPi - \mbPi_{0}} + \text{penalty term} $. \emph{(ii)} When learning actions or states, step 6 can be solved by traversing all the possible $ \mathbf{T} $. \emph{(iii)} In step 6 of Algorithm \ref{alg:online-iop}, solving \eqref{eq. GIRL objective} - \eqref{eq. GIRL constraint 5} can be quite efficient since it is a linear program. 
\end{remark}

When the state space $ \mcS $ is infinite, a popular method is to discretize $ \mcS $ into a finite state space $ \tilde{\mcS} $ \cite{Bertsekas05}. Then, we use the model in Sections \ref{sec: GIRL in finite state spaces} and \ref{section:custimize girl} to learn action, state and transition probability matrices. Nevertheless, it is hard to use GIRL to directly learn a reward function due to the large size of $ \tilde{\mcS} $. Therefore, instead of learning the reward function directly, we approximate it as $ R(s) = \sum_{i=1}^{d} \theta_i \phi_i(s) $, where $ \{\phi_1, \ldots, \phi_d\} $ are fixed and bounded basis functions, and $ \{\theta_1, \ldots, \theta_d\} $ are the parameters we want to learn. Replacing $ R $ by its linear approximation, GIRL gives us an estimation of parameters $ \{\theta_1, \ldots, \theta_d\} $.

\section{Experiments}
	\label{sec: experiments}
	In this section, we illustrate the performance of Algorithm \ref{alg:online-iop} in a discrete grid world problem and a continuous grid world problem. Results show that our algorithm solves GIRL quite efficiently and yields good inferences of the unknown components of an MDP and the underlying optimal policy.
	
	\textbf{Discrete grid world}\;\;
	In the first experiment, we consider a $ 5\times 5 $ grid world, where the agent starts from the lower-left grid square, and intends to move to the (absorbing) upper-right grid square to receive a reward of 1. The action space is moving in the four compass directions. After a decision is made, noise leads to a 30\% chance of moving in a random direction. Figure \ref{fig: grid5, optimal policy} shows an optimal policy $ \pi $ and Figure \ref{fig: grid5, true reward} is the true reward function. We enumerate the states left-to-right and then bottom-to-top, with the initial state labeled as `1' and the absorbing state labeled as `25'.
	
	The observed policy $ \pi_{0} $ is generated in the following way. Note that no state takes action `$ \downarrow $' under the optimal policy in Figure \ref{fig: grid5, optimal policy}. Thus, we can add noise to the optimal policy by forcing $ \pi(\downarrow | s) = 1 $ for some $ s $, which is called a \textbf{noisy state}. Specifically, when there are $ n $ noisy states, the actions taken in the first $ n $ states are `$ \downarrow $'. 
	
	We study five different learning tasks: (1) reward function, (2) transition probability matrices, (3)  state space, (4) action space, and (5) reward + action space. 
	In tasks (1), we add $ \norm[0]{R} $ to the objective function of GIRL as the penalty term. In task (2), we consider the case where in the transition probability matrix under action `$ \rightarrow $', the first two elements of the row corresponding to state 2 are unknown. In task (3), we consider a grid world with one blocked square as shown in Figure \ref{fig: grid5, block}. The blocked square and its neighbor on the left are covered so that we cannot locate the blocked one. Our goal is to determine which square is the blocked one given the observed policy. In task (4), given $ \{\uparrow,\leftarrow,\downarrow\} \in \mcA $, we want to figure out the unobservable action, whose candidates are $ \{\rightarrow,\nearrow\} $. Task (5) combines tasks (1) and (4). 
	
	\newcommand{\xmark}{\ding{55}}%
	
	To measure the performance of GIRL, we consider two criteria. One is the average distance between the true value of the unobservable or partially observable component and its estimation, which is shown in Table \ref{table: criteria}. $ d_{H} $ stands for the Hamming distance. The other one is whether an optimal policy is recovered. If yes, we denote it by \checkmark, otherwise \xmark. The results are shown in Table \ref{table. grid5}. For the first criterion, the average distances are all 0 except for the case of task (1) and (5) with 4 noisy states, indicating that GIRL performs quite well in handling these tasks. For the second criterion, GIRL succeeds in recovering an optimal policy in every learning task. Furthermore, Figure \ref{fig: grid5, noise1} show the estimated reward function in task (1) with 4 noisy states. 
	
	\begin{figure}[ht]
		\centering
		\begin{subfigure}[ht]{0.20\textwidth}
			\includegraphics[width=1\textwidth]{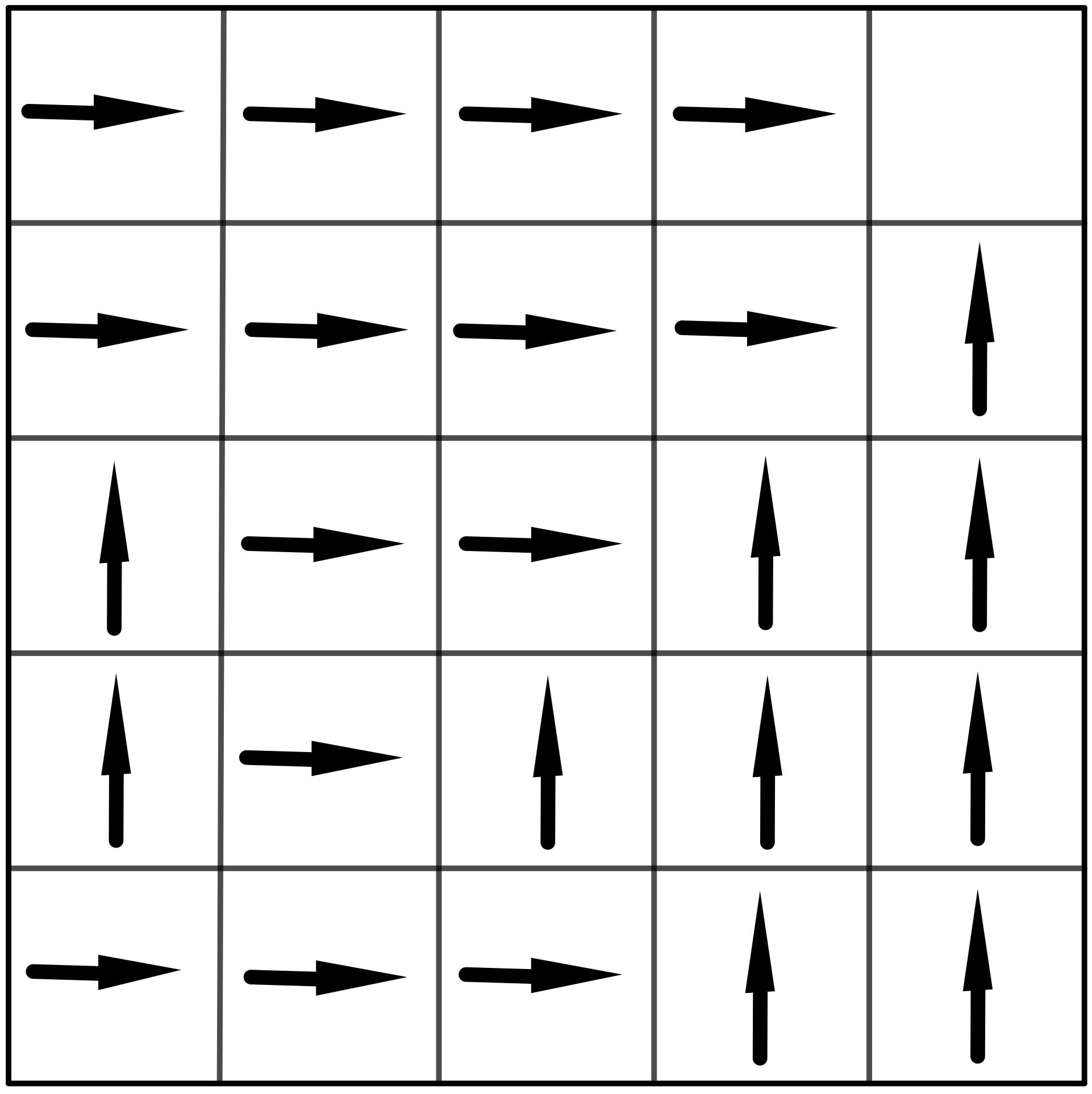}
			\caption{}
			\label{fig: grid5, optimal policy}
		\end{subfigure}\;\;
		\begin{subfigure}[ht]{0.20\textwidth}
			\includegraphics[width=1\textwidth]{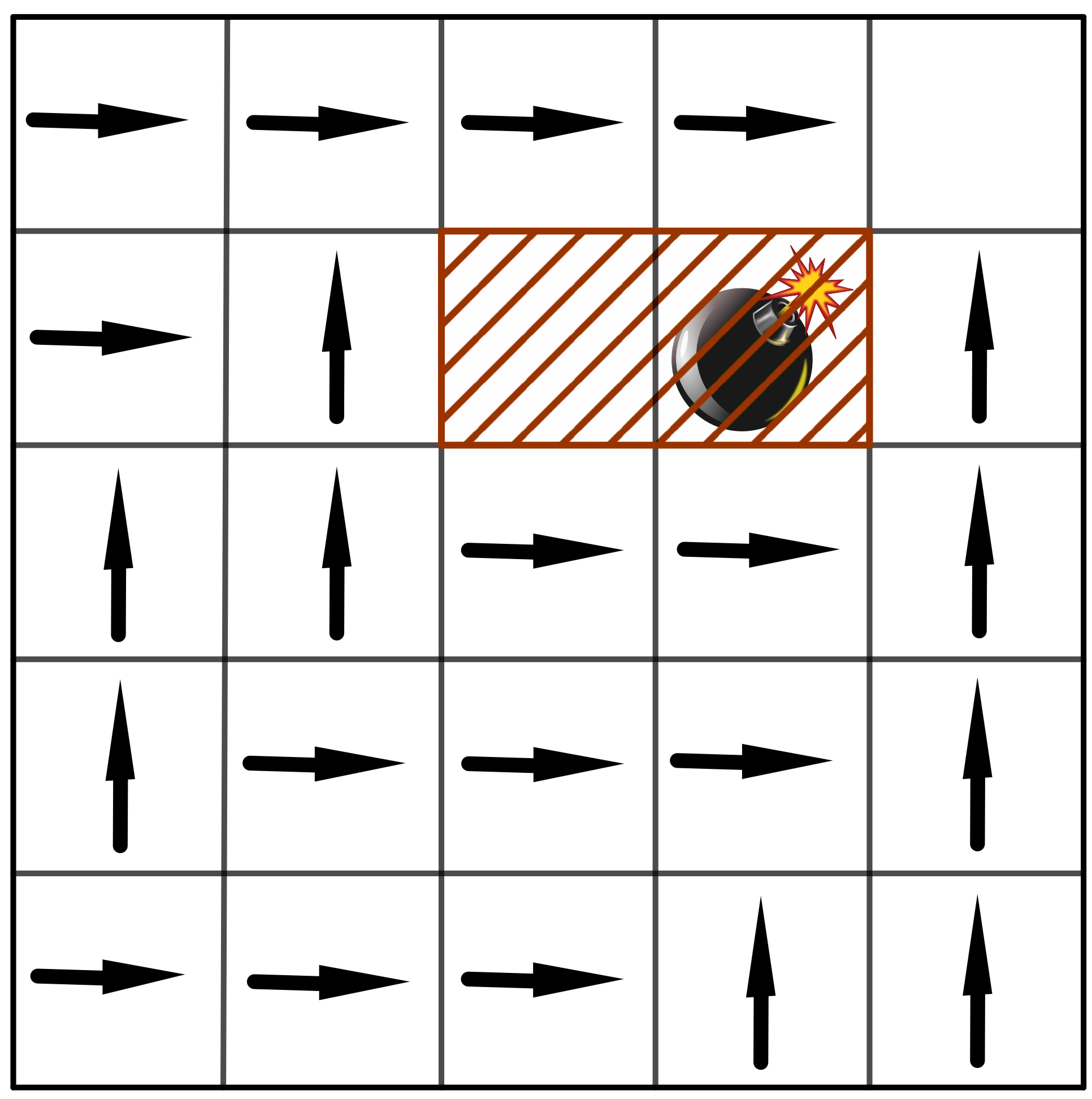}
			\caption{}
			\label{fig: grid5, block}
		\end{subfigure}\;\;
		\begin{subfigure}[ht]{0.265\textwidth}
			\includegraphics[width=1\textwidth]{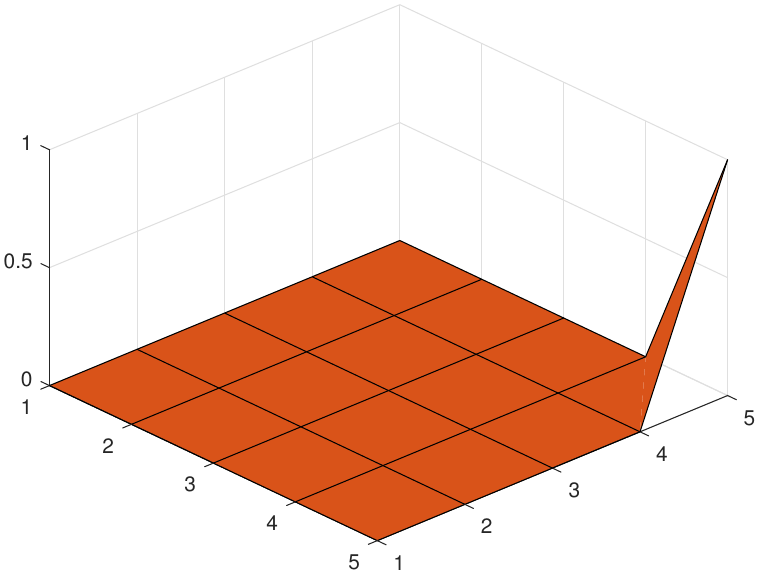}
			\caption{}
			\label{fig: grid5, true reward}
		\end{subfigure}\;\;
		\begin{subfigure}[ht]{0.265\textwidth}
			\includegraphics[width=1\textwidth]{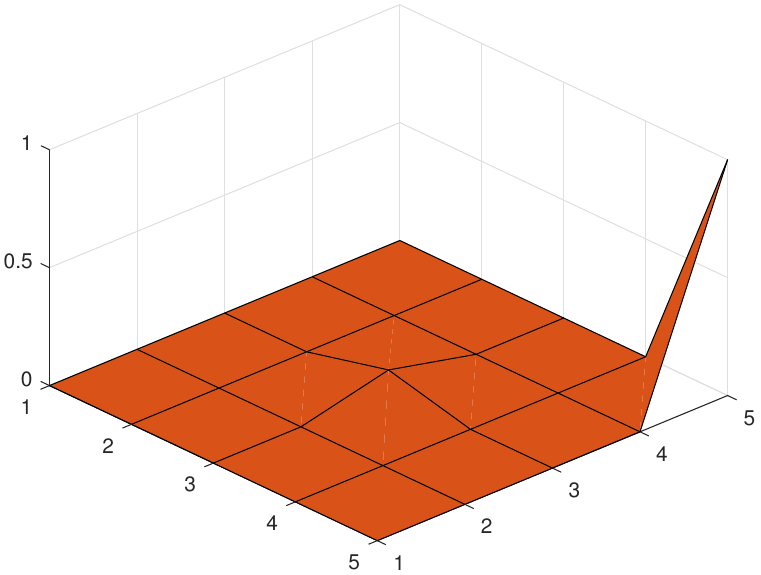}
			\caption{}
			\label{fig: grid5, noise1}
		\end{subfigure}
		\caption{Learning the reward function for discrete grid world. (a) The $ 5\times 5 $ grid world with optimal policy. (b) Blocked grid world. (c) True reward. (d) Estimated reward.}
		\label{fig: grid5, optimal}
	\end{figure}

     \begin{table}[ht]
     	\centering
     	\caption{The first Criterion to measure the performance of GIRL}
     	\label{table: criteria}
     	\begin{tabular}{@{}ccccc@{}}
     		\toprule
     		Task     & Reward & TransitionProb Matrix & State & Action  \\ \midrule
     		Criterion 1 &   $ \norm[1]{R - \hat{R}}/|\mcS| $     &      $ \norm[1]{P - \hat{P}}/|\mcS|^2 $                &    $ d_{H}(\mcS, \hat{\mcS})/|\mcS| $   &        $ d_{H}(\mcA, \hat{\mcA})/|\mcA| $           \\ \bottomrule
     	\end{tabular}
     \end{table}
 
     \begin{table}[ht]
     	\centering
     	\caption{Performance of the algorithm for different learning tasks}
     	\label{table. grid5}
	\begin{tabular}{ccccccccccc}
		\hline
		Number of noisy states        & \multicolumn{2}{c}{0}         & \multicolumn{2}{c}{1}         & \multicolumn{2}{c}{2}         & \multicolumn{2}{c}{3}         & \multicolumn{2}{c}{4}              \\ \cline{2-11} 
		Criterion                     & 1 & 2                         & 1 & 2                         & 1 & 2                         & 1 & 2                         & 1      & 2                         \\ \hline
		Reward                        & 0 & \checkmark & 0 & \checkmark & 0 & \checkmark & 0 & \checkmark & 0.0035 & \checkmark \\
		Transition probability matrix & 0 & \checkmark & 0 & \checkmark & 0 & \checkmark & 0 & \checkmark & 0      & \checkmark \\
		State                         & 0 & \checkmark & 0 & \checkmark & 0 & \checkmark & 0 & \checkmark & 0      & \checkmark \\
		Action                        & 0 & \checkmark & 0 & \checkmark & 0 & \checkmark & 0 & \checkmark & 0      & \checkmark \\
		Reward + Action               & 0 & \checkmark & 0 & \checkmark & 0 & \checkmark & 0 & \checkmark & 0.0035 & \checkmark \\ \hline
	\end{tabular}
\end{table}

	\textbf{Continuous grid world}\;\;
	The next experiment is run on a continuous grid world. The state space is $ [0,1] \times [0,1] $, the action space is to move the agent 0.2 in the four compass directions. After a decision is made, an uniform noise in $ [-0.1,0.1] \times [-0.1,0.1] $ is added. The noise is truncated if necessary to keep the agent within the grid world. The true reward is 1 in the upper right square $ [0.8,1] \times [0.8,1] $, and 0 everywhere else, and $ \gamma=0.9 $. 
	
	To discretize the state space, we consider three cases: a $ 10\times10 $ grid world, a $ 20\times20 $ grid world, and a $ 30\times30 $ grid world. For each discretization, the state in each period is approximated by the square that covers the exact state. To generate the transition probability matrices, we simulate the agent's movement in each state under each action according to the noise rule 1,000 times.
	
	The observed policy is generated in the same way as the previous experiment. We focus on learning the reward function, which is approximated by a linear combination of some evenly spaced multivariate Gaussian-shaped basis functions. The number $ d $ of basis functions we use is adjusted according to the discretization: $ d=25 $ in the $ 10\times10 $ grid world, $ d=100 $ in the $ 20\times20 $ grid world, $ d=225 $ in the $ 30\times30 $ grid world. 
	
\begin{figure*}
		\centering
		\begin{subfigure}[ht]{0.3\textwidth}
			\includegraphics[width=1\textwidth]{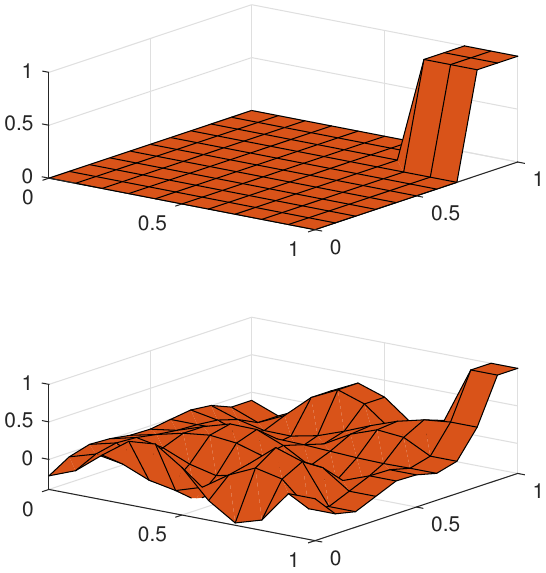}
			\caption{$ 10 \times 10 $}
			\label{fig: grid_continuous10, true reward}
		\end{subfigure} 
		\begin{subfigure}[ht]{0.3\textwidth}
			\includegraphics[width=1\textwidth]{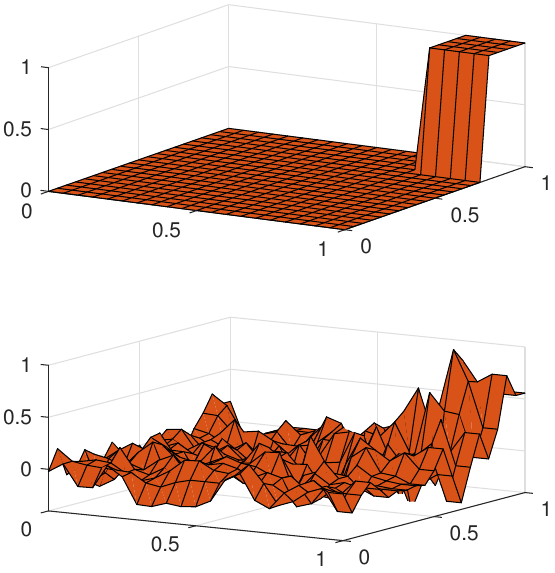}
			\caption{$ 20 \times 20 $}
			\label{fig: grid_continuous10, noise1}
		\end{subfigure} 
	    \begin{subfigure}[ht]{0.3\textwidth}
	    	\includegraphics[width=1\textwidth]{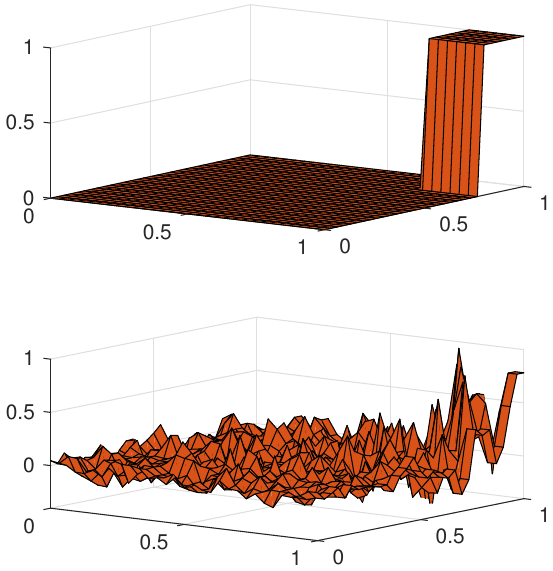}
	    	\caption{$ 30 \times 30 $}
	    	\label{fig: grid_continuous15, true reward}
	    \end{subfigure}
		\caption{Learning the reward function for continuous grid world. Number of noisy states is set to be 2. Figures at the top are the true rewards under the corresponding discretization. Figures on the bottom are the estimated rewards. (a) $ 10 \times 10 $ discretization of the state space. (b) $ 20 \times 20 $ discretization of the state space. (c) $ 30 \times 30 $ discretization of the state space.}
		\label{fig: grid_continuous}
	\end{figure*}

	Table \ref{table: grid_continuous} shows the computational results of different discretization of the state space and different numbers of noisy states. We use the average distance between the true reward $ R $ and the estimation $ \hat{R} $ to measure the performance of GIRL. As shown in the table, for fixed discretization of the state space, $  \norm[1]{R - \hat{R}}/|\mcS| $ is monotone increasing with the number of noisy states. This makes great sense because both the reward function and the underlying optimal policy become less likely to be identifiable when the observed policy has more noises. Thus, with more noises in the observed policy, our algorithm tends to return an estimated policy matrix that is closest to the observed policy matrix, while the associated reward function might not be close to the true reward function. Moreover, for fixed number of noisy states, we can see from the table that the estimation is better with finer discretization of the state space. The main reason is that finer discretization will increase the size of the observed policy matrix, and thus reduce the relative noise level. To further illustrate the performance of the proposed algorithm, a plot of the true reward functions and the estimated reward functions of the three different discretization with two noisy states is given in Figure \ref{fig: grid_continuous}. All of the three estimated reward functions match the structures of the true reward functions, indicating that our algorithm indeed can be used to learn the reward functions. 
	
	\begin{table}[ht]
		\centering
		\caption{Average distance between the true reward and the estimation}
		\label{table: grid_continuous}
		\begin{tabular}{cccccc}
			\hline
			Number of noisy states & 0      & 1      & 2      & 3      & 4      \\ \hline
			10 $\times$ 10 discretization    & 0.1563 & 0.1563 & 0.1563 & 0.1563 & 0.1571 \\
			20 $\times$ 20 discretization    & 0.0955 & 0.1125 & 0.1169 & 0.1285 & 0.1383 \\
			30 $\times$ 30 discretization    & 0.0799 & 0.0915 & 0.0962 & 0.1005 & 0.1074 \\ \hline
		\end{tabular}
	\end{table}
	
\vspace{-5pt}	
\section{Conclusions and Future Work}

    In this paper, we address the problem of generalized inverse reinforcement learning. We propose the mathematical formulation for GIRL and develop an algorithm that can solve moderate-sized problem. Experiments on a variety of learning tasks with finite or infinite state space show that our approach yields good inferences of the unknown components of an MDP, and recovers optimal polices for these tasks. Similar to IRL, the unidentifiability issue exists in GIRL as well. Several exciting works have been proposed to address this issue in IRL \cite{ramachandran2007bayesian,ziebart2008maximum,finn2016guided,NIPS2017_6778}. Thus, an interesting direction of future work is to incorporate their resolutions to the unidentifiability into GIRL.
	\bibliographystyle{unsrt}
	\bibliography{reference}
\end{document}